\documentclass[runningheads]{llncs}
\usepackage[T1]{fontenc}
\usepackage{graphicx}
\usepackage{verbatim}
\usepackage{enumitem}
\usepackage{cite}
\usepackage{amsmath}
\usepackage{xcolor}
\usepackage{booktabs}
\usepackage{amssymb}
\usepackage{hyperref}

\begin{document}
\title{CalcSeg: Confidence-aware 3D Latent Context Curriculum Learning For Myocardial Scar Segmentation From Single-Stack LGE-CMRs}

\author{Nivetha Jayakumar\inst{1} \and
Hannah Kim\inst{1} \and
Amit R. Patel\inst{2} \and
Miaomiao Zhang\inst{1}}
%index{Jayakumar, Nivetha}
%index{Kim, Hannah}
%index{Patel, Amit}
%index{Zhang, Miaomiao}

%
\authorrunning{N. Jayakumar et al.}

\institute{School of Engineering and Applied Science, University of Virginia, USA\\
\and
School of Medicine, University of Virginia, USA \\
\email{vfb8zv@virginia.edu}}

\maketitle          
\begin{abstract}
Myocardial scar segmentation from single-stack late gadolinium–enhanced cardiac magnetic resonance (LGE-CMR) imaging has been a longstanding and clinically important challenge, particularly in the presence of low tissue contrast, diffuse, and small scar regions. These challenges are further intensified by the limited availability of 3D spatial context. This paper presents {\em CalcSeg}, a \underline{C}onfidence-\underline{a}ware \underline{l}atent context \underline{c}urriculum learning framework that leverages fused 3D feature representations from single-stack 2D LGE-CMR images for robust scar segmentation. Specifically, we introduce a dynamic semi-supervised curriculum learning strategy that progressively expands training from easier to more challenging scar cases using a learned confidence-aware scoring function. Such a function integrates errors in the predicted scar maps with quantified epistemic uncertainty and scar burden estimation to automatically assess sample difficulty without requiring manual labels. To compensate for the limited spatial context in single-stack acquisitions, we then develop a latent slice-wise self-attention to capture inter-slice dependencies and infer 3D spatial representations from sparse 2D inputs. We evaluate CalcSeg on multi-center clinical LGE-CMR datasets and benchmark against existing scar segmentation networks. Experimental results show that CalcSeg consistently outperforms all competing methods, particularly with substantial improvements on clinically challenging cases. Our code is released on \href{https://github.com/vfb8zv/CalcSeg}{Github}


\end{abstract}

\section{Introduction}
Myocardial scar segmentation from a single-stack LGE-CMR plays an important role in the detection and quantification of myocardial fibrosis and infarction in patients with heart diseases~\cite{kwong2011measuring}. The presence and extent of scar burden have shown to strongly predict adverse cardiovascular outcomes, including sudden cardiac death, heart failure hospitalization, and mortality in ischemic and non-ischemic cardiomyopathies~\cite{kellman2012cardiac,coleman2017prognostic}. Despite its clinical significance, manual scar delineation remains labor-intensive and often subject to inter-observer variability~\cite{rajchl2015comparison}. A fully automated and robust method is needed in routine clinical workflows to provide accurate and reproducible detection of myocardial scar. 

Recent deep learning has achieved superior performance in myocardial scar segmentation from LGE-CMR images~\cite{ronneberger2015u,xing2023joint,tavakoli2025scarnet,zou2025scarelastic}.
However, most existing approaches operate on 2D slice-wise inputs and are primarily developed on cases with clear contrast and well-defined scar regions~\cite{karim2016evaluation}, or require multi-sequence input to achieve high performance~\cite{qiu2023myops}. As a result, segmentation performance remains challenging in clinically difficult scenarios with low-contrast enhancement, diffuse, and small scars~\cite{jayakumar2025deep,hoh2024impact}. Prior efforts have attempted to address these issues with various strategies, including leveraging large foundation-models with boundary-robust metrics and data augmentation to increase the model's robustness to challenging samples~\cite{moafi2025robust}. Other research groups utilize generative models to improve low contrast samples before performing segmentation tasks~\cite{hasny2024myocardial}. Later, curriculum learning (CL) gained increasing attention in segmentation, as it offers a principled strategy to improve model performance by structuring training from high-confidence, reliable samples to more challenging or lower-confidence cases~\cite{bengio2009curriculum,zhou2020robust}. However, existing CL typically rely on static predefined difficulty definitions, sample reweighting, or local-to-global schedules~\cite{yang2025progressive,zheng2024curriculum,jayakumar2026curriculum}. They do not explicitly model subject-level variability or pathology-specific progression during training, which leave diffuse, heterogeneous, or uncertain cases insufficiently emphasized during learning~\cite{cui2022deep}. While further works introduce partial inter-slice information from single-stack CMRs~\cite{crozieruncertainty,zhang2020cascaded}, they do not model a complete 3D anatomical relationship to fully capture myocardial scars.  

In this paper, we present CalcSeg - a confidence-aware curriculum learning for LGE-CMR scar segmentation that explicitly accounts for sample difficulty with learned 3D latent context in single-stack data. Specifically, we introduce a novel, dynamic, and sample-specific confidence score derived from clinically relevant segmentation error metrics. This confidence measure induces a continuation schedule in the training sample space, progressively expanding the effective training set from high-confidence (easy) cases to more challenging ones. By smoothly modulating the optimization landscape across stages, CalcSeg transitions from stable coarse learning to fine-grained refinement with improved robustness. 
Additionally, we develop a latent fusion module with slice-wise self-attention to capture subject-level inter-slice correlations across the full LGE stack. This enables effective modeling of 3D contextual dependencies while preserving slice-specific features; hence enhancing consistency and anatomical plausibility in myocardial scar segmentation. Our contributions are threefold:
\begin{enumerate}[label=(\roman*.)]
    \item Develop an semi-supervised confidence-aware curriculum learning of sample space that progressively expands training from easy to difficult landscape. 
    \item Introduce a latent 3D context formulation from single stack LGE-CMR with slice-wise self-attention to enforce subject-level anatomical consistency.
    \item Experimental results show that our model achieves superior performance over existing segmentation models across multi-site data with challenging cardiomyopathy cases.
\end{enumerate}

\section{Background: Curriculum Learning Scheme}
This section briefly reviews the training paradigm of CL~\cite{bengio2009curriculum}, which was employed to improve deep network optimization by presenting training samples in a meaningful order (i.e., progressing from easy to hard examples)~\cite{krueger2009flexible}. Inspired by human learning processes and continuation optimization strategies~\cite{pathak2025principled}, CL has been shown to stabilize and accelerate convergence of the network training process. In contrast to conventional training schemes that implicitly assume a fixed notion of sample difficulty and rely on random sampling from the input data distribution or loss landscape~\cite{tavakoli2025scarnet}, CL explicitly controls task complexity during training. By gradually increasing sample difficulty, the model is encouraged to first capture coarse and dominant structures before learning more challenging, ambiguous, or rare patterns~\cite{gao2025principled}.

Consider a training dataset of $N$ samples, $\mathcal{D} = \{(x_i, y_i)\}_{i=1}^{N}$, with network parameters, $\theta$. CL can be formulated as training over a sequence of progressively expanded subsets, $S_1 \subset \cdots S_k \subset \cdots \subset S_t = \mathcal{D}, k \in [1, t]$, where $S_k$, includes samples up to a predefined difficulty level. Given a network loss function, $L(\theta; x_i, y_i)$, the optimization of CL at each stage $k$ can be formulated as: 
\begin{equation}
\min_{\theta} \sum_{(x_i,y_i)\in S_k} L(\theta; x_i, y_i).
\label{eq:intro_loss}
\end{equation}

The CL scheme can be interpreted as a continuation optimization strategy over the complete objective in Eq.~\eqref{eq:intro_loss}~\cite{pathak2025principled}. From this point of view, the loss landscape is progressively reshaped as increasingly difficult samples are incorporated into training. Specifically, the early stages optimize a smoother objective constructed from high-confidence (easy) samples, yielding stable gradients and well-conditioned feature representations. As training proceeds, lower-confidence and more challenging samples are gradually introduced; thereby refining the solution toward the true objective. By initializing each stage with parameters learned from the previous stage, the CL optimization follows a continuous solution trajectory with improved stability~\cite{krueger2009flexible,pathak2025principled}. 

\section{Our Method} 
\begin{figure}[!hb]
\centering
\centerline{\includegraphics[width=0.99\linewidth]{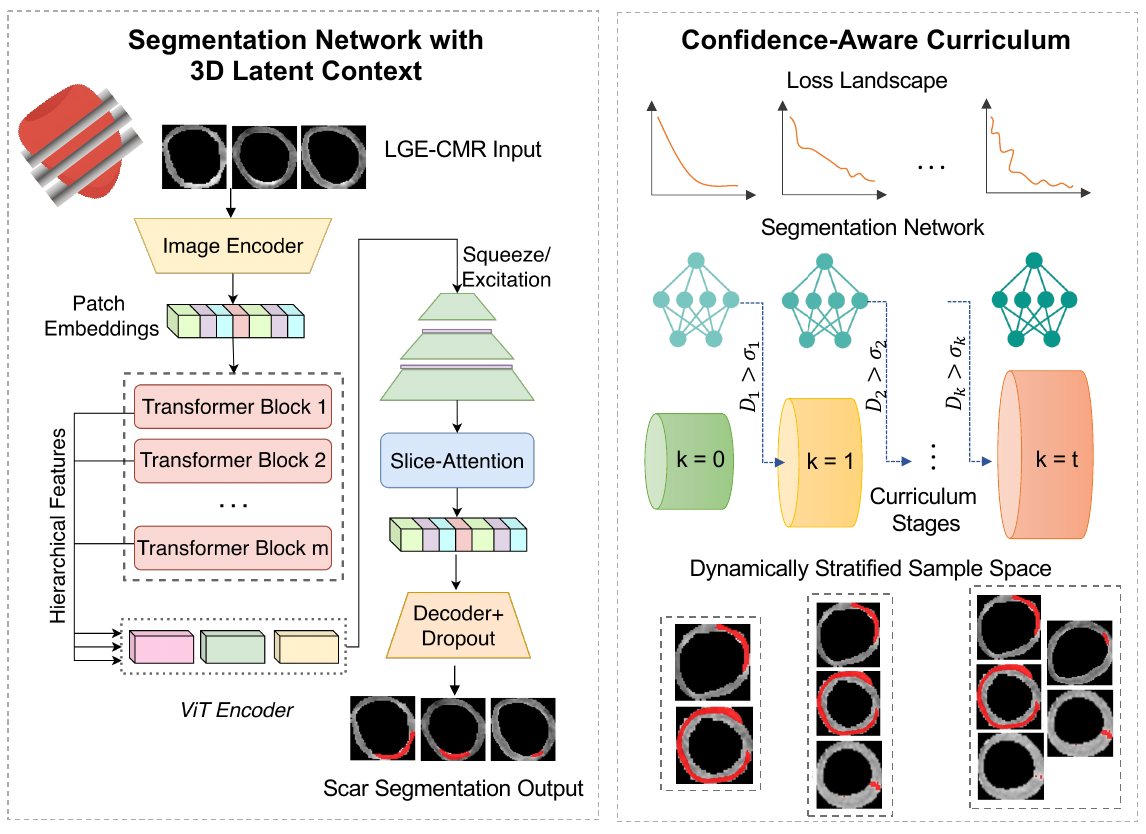}}
\caption{Curriculum-guided training framework for LGE segmentation.}
%\vspace{-2mm}
\label{fig:main_arch}
\end{figure}

This section outlines the details of CalcSeg, a confidence-aware CL framework that performs automatic scar delineation with improved precision, particularly in challenging cases. Our model features two key components: (i) a latent 3D context formulation with slice-wise self-attention for anatomical consistency, and (ii) a confidence-aware curriculum that progressively expands training from confident to ambiguous samples. An overall architecture is shown in Fig.~\ref{fig:main_arch}. \\

\noindent\textbf{Problem Setting.} Let $\mathcal{D} = \{(x_i, y_i)\}_{i=1}^N$ be the dataset consisting of LGE images $x_i$ and its corresponding ground-truth scar segmentation labels $y_i$. We design a deep network $F_\theta$, with parameters $\theta$, to predict the segmentation as $\hat{y(\theta)_i} = F_\theta(x_i)$. 

\noindent\textbf{3D Latent Context Modeling.} We first introduce a 3D latent attention module (as shown on the left panel of Fig.~\ref{fig:main_arch}) that explicitly models subject-level 3D representations inferred from slice-wise 2D features, $z$; thereby improving segmentation across sparse LGE stacks. In contrast to previous approaches that process individual 2D slices~\cite{tavakoli2025scarnet}, our model learns hierarchical inter-slice context in the latent space. More specifically, we develop an autoencoder architecture that integrates a Transformer-based encoder~\cite{dosovitskiy2020image,ma2024segment} with a multi-scale convolutional decoder enhanced by a  {\em disentangled slice-wise self-attention}~\cite{arnab2021vivit}. This module is designed to effectively capture subject-level anatomical context by performing explicit 2D-to-3D latent feature fusion on the encoded latent features, $z$. In order to preserve slice order information, fixed sinusoidal positional embeddings are integrated along the slice dimension prior to attention modeling.

In practice, we implement this slice-wise attention as a single-head self-attention, where queries ($Q$), keys ($K$), and values ($V$) are first obtained as linear projections of $z$, followed by the attention score computation as $A = \text{softmax}\!\left(\frac{QK^\top}{\sqrt{C}}\right) V$. Here $C$ denotes the projection dimension. To support variable stack sizes, a binary mask suppresses padded slices by setting their attention scores to $-\infty$ before the softmax. The resulting slice-aware latent features encode subject-level anatomical context and are passed to the decoder for final segmentation. To reduce model complexity, we remove the auxiliary UNet branch used in prior work~\cite{tavakoli2025scarnet} and instead introduce a custom residual decoder that refines multi-scale features. \\
\noindent\textbf{Confidence-aware Curriculum Learning.} In contrast to conventional CL strategies that rely on predefined or static difficulty scores for each training sample~\cite{gao2025principled}, we introduce a novel semi-supervised, model-adaptive difficulty formulation grounded in the network’s own training dynamics and predictive behavior. Instead of assuming difficulty as an intrinsic property of the data, our approach dynamically quantifies it based on how the model performs and how confidently it makes predictions. For each data point, we define the difficulty score using two complementary components:
\begin{enumerate}[label=(\roman*)]
    \item {\em Predictive Inaccuracy}. We measure task-relevant prediction error using two clinically and geometrically meaningful metrics: (a) the Dice Similarity Coefficient, $\rho$, which evaluates volumetric overlap between the predicted segmentation and ground truth~\cite{dice1945measures}, and (b) the percentage error in scar burden, $v$, a clinically established measure quantifying total myocardial scar volume~\cite{jayakumar2025deep}. These metrics ensure that difficulty reflects both spatial accuracy and clinically significant volumetric deviation.
    \item {\em Predictive Uncertainty}. We quantify epistemic uncertainty by measuring the variance of model outputs under stochastic forward passes. In this paper, we estimate uncertainty, $\epsilon$, via Monte Carlo Dropout~\cite{gal2016dropout} applied to the final network layer, capturing the model’s confidence in its predictions. Similarly, other uncertainty quantification metrics~\cite{he2025survey} can also be applied. 
\end{enumerate}

We compute these metrics during training and update them across 
CL stages, $k\in[0,t]$, to determine subsets $\{S_k\}_{k=0}^t$. For each sample $i$, the 
difficulty components are defined as $D_i=[\gamma_1\rho_i,\gamma_2 v_i,\gamma_3 \epsilon_i],$ where $\{\gamma_j\}_{j=1}^3$ are weighting parameters. At stage $k$, the subset is determined by a confidence scoring function as
\begin{equation}
    \alpha_{i,k} (D_i)=
\begin{cases}
1, & \text{if } \gamma_1\rho_i<\sigma_k^{(1)} | \gamma_2v_i>\sigma_k^{(2)} | \gamma_3\epsilon_i>\sigma_k^{(3)}, \\[2pt]
0, & \text{otherwise}.
\end{cases}
\end{equation}
where $\sigma_k$ denotes stage-dependent thresholds that are calculated automatically based on the median of the scores at each stage. This assigns a value of $1$ to difficult samples if any of the components exceeds its threshold. \\
\noindent {\bf Sample-Space Continuation.} To implement the proposed curriculum, we adopt a homotopy-based optimization~\cite{lin2023continuation} that progressively transforms the learning objective from an initially simplified problem defined by easier LGE cases to the full learning objective involving increasingly difficult examples. The total loss at any stage $k$ is given as: 

\begin{center}
${\mathcal{L}}_{\theta,k} = \frac{1}{\hat{N}_k}\sum_{i=1}^{\hat{N}_k} [- {\beta_k}  Y_i \cdot \hat{Y}_i'^{\delta_k}  \log(\hat{Y}_i) - (1 - \beta_k) Y'_i \cdot \hat{Y_i}^{\delta_k}  \log( \hat{Y}'_i)] +$ \\
$
1 - \Biggl(
w_{\text{fg},k} \frac{2 \sum_{i \in M} Y_i \hat{Y}_i}{\sum_{i \in M} Y_i + \sum_{i \in M} \hat{Y}_i} 
+ 
(1-w_{\text{fg},k}) \frac{2 \sum_{i \in M} Y_i' \hat{Y}_i'}{\sum_{i \in M} Y_i' + \sum_{i \in M} \hat{Y}_i'}
\Biggr),
$
\end{center}
where $\hat{N}_k \triangleq size(S_k)$, $y' \triangleq (1-y)$. The $\mathcal{L}_\theta$ represents a pixel-wise segmentation loss, which is a weighted sum of the focal and foreground-weighted Dice loss minimized over the stages (as given in Eq.~\ref{eq:intro_loss}). The weighting parameter, $w_{\text{fg},k}$, balances the relative contributions of foreground and background overlap, while $\beta_k$ and $\delta_k$ control the weightage on positive-negative cases and hard-to-classify pixels at each stage $k$.

\section{Experimental Results}

This section describes the dataset used in our experiments, parameter setting of the proposed framework, evaluation metrics used to quantify segmentation performance, and comparison against SOTA LGE segmentation methods. \\

\noindent\textbf{Dataset.} We evaluate the proposed method on LGE-CMR datasets from four sites~\cite{wang2022ai,paudel2026cardiac} and two segmentation challenges, including the MICCAI 2012 Left Ventricular (LV) Infarct Challenge~\cite{karim2016evaluation} and the EMIDEC 2020 challenge~\cite{lalande2020emidec}, comprising $976$ patients diagnosed with ischemic and non-ischemic cardiomyopathies and without fibrosis, where each subject has between 4-18 slices. Manual annotations of the LV and scar regions were provided by clinical experts. All slices were masked to the LV region using these contours, and further cropped and resized to $224^2$. The dataset's train/validation/test split is 7:1:2.

\noindent\textbf{Parameter setting.} We load the transformer encoder's weights from MedSAM~\cite{ma2024segment} and pretrain the autoencoder on stacks of 2D LGE slices, through random mini-batching and the Adam optimizer with a learning rate of $1e-4$. This model serves as a baseline to start the curriculum stages. The foreground weighting for the dice loss, $w_{fg}$ ranges from $0.6$ to $0.8$, $\beta$ of the focal loss from $0.25$ to $0.35$, and $\delta$ from $1.25$ to $2$ based on the difficulty score $D_i$. Although these parameters can be approximated from $D_i$, we use fixed intervals for ease of optimization. The dropout probability of the decoder is $0.3$. The loss terms are weighted by the inverse of their order of magnitude to ensure uniform numerical range. We obtain the mean standard deviation of $100$ MC Dropout samples per subject to quantify the uncertainty. 
All experiments were carried out on NVIDIA A40 and V100 GPUs. 

\noindent\textbf{Evaluation against SOTA methods.} We quantify the segmentation performance using Dice Similarity Coefficient and the percentage error in scar volume computed over the myocardium for the complete independent test set, and $6$ known challenging cases explicitly flagged by clinical experts. These metrics quantify overlap accuracy and clinically relevant volumetric error, respectively. The proposed method is compared against $5$ state-of-the-art baselines, including TransUNet~\cite{chen2021transunet}, AttentionUNet~\cite{oktay2018attention}, UNETR~\cite{hatamizadeh2022unetr}, ScarNet~\cite{tavakoli2025scarnet} and the current SOTA ScarNet with supervised CL, where we leverage expert-provided difficulty labels for curriculum-guided training. 

\begin{figure}[!t]
\centering
\centerline{\includegraphics[width=1.0\linewidth]{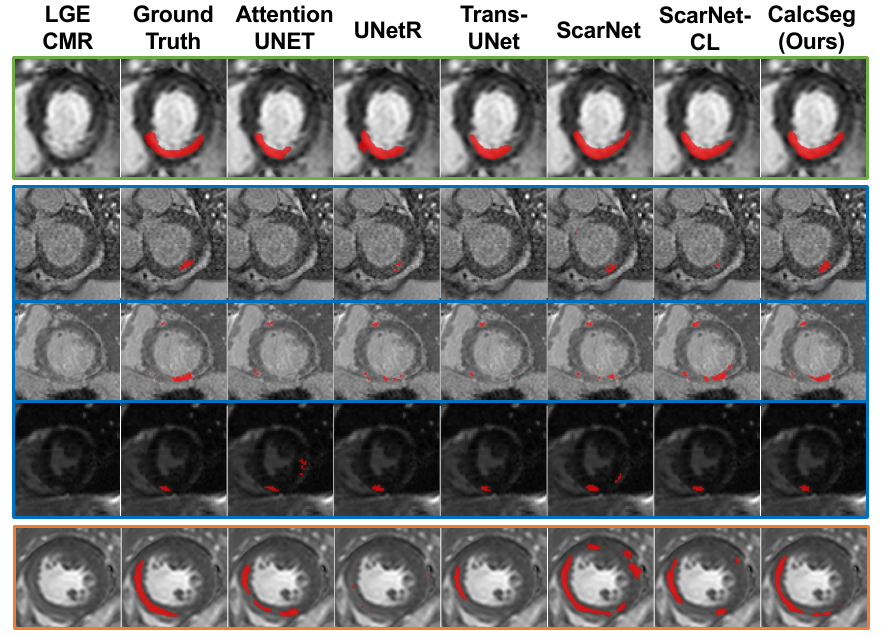}}
\caption{Qualitative LGE segmentation results: examples of clinically determined high (green) and low-confidence (red) cases vs. diffusive scar or low-contrast cases without a clinical confidence label (blue).}
\label{fig:main_res}
\end{figure}

\noindent\textbf{Results.} Fig.~\ref{fig:main_res} shows qualitative results of our method compared to the baselines. CalcSeg achieves higher overlap with the ground truth scar as compared to the baselines with minimal false positives.  

\begin{figure}[!htbp]
\centering
\centerline{\includegraphics[width=1.0\linewidth]{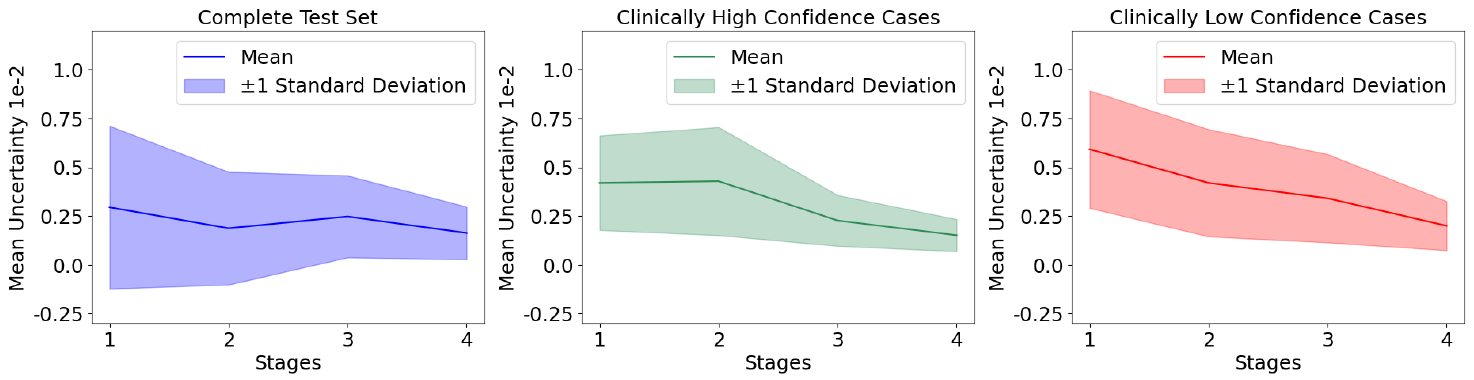}}
\caption{Uncertainty quantified as the standard deviation of the Monte Carlo samples at the end of each stage of SCL, where it progressively decreases with the stages.}
\label{fig:unc_res}
\end{figure}

Fig.~\ref{fig:unc_res} shows the quantified uncertainty at the end of each stage decreasing as the stages progress. The overall mean (left, Fig.3) appears relatively stable since high-confidence cases dominate the test set (ratio 22:6). In contrast, when stratified by difficulty (right, Fig.3), uncertainty decreases substantially across stages for the clinically challenging subgroup.

Tab.~\ref{tab:sota} summarizes the evaluation metrics, demonstrating that our method achieves higher overall performance. Our approach achieves improved scar burden estimation and higher myocardial agreement including healthy tissue, indicating that our framework effectively balances sensitivity and specificity. \\

\begin{table}[t]
\centering
\setlength{\tabcolsep}{6pt}
\caption{Quantitative comparison of LGE segmentation performance across SOTA baselines, reporting Myocardial Scar Dice (Dice), Low Confidence samples' Dice ($\text{Dice}_\text{low}$), Scar Error (SE) and Low Confidence samples' Scar Error ($\text{SE}_\text{low}$).}
\label{tab:sota}
\begin{tabular}{lcccc}
\toprule
Model & Dice $\uparrow$ & $\text{Dice}_\text{low}$ $\uparrow$ & SE (\%) $\downarrow$ &$ \text{SE}_\text{low}$ (\%) $\downarrow$ \\
\midrule
TransUNet & 0.617 $\pm$ 0.24 & 0.460 $\pm$ 0.22  & 48.48 & 46.09 \\
AttentionUNet & 0.601 $\pm$ 0.25 & 0.454 $\pm$ 0.27 & 49.76 & 50.88 \\
UNETR & 0.524 $\pm$ 0.25 & 0.404 $\pm$ 0.26 & 57.22 & 45.95 \\
ScarNet & 0.615 $\pm$ 0.29 & 0.448 $\pm$ 0.21 & 46.11 & 49.97 \\
ScarNet (Sup. CL) & 0.644 $\pm$ 0.29 & 0.574 $\pm$ 0.23 & 45.07 & 48.34\\
\midrule
\textbf{CalcSeg (Ours)} & \textbf{0.677 $\pm$ 0.24} & \textbf{0.644 $\pm$ 0.22} & \textbf{38.88} & \textbf{35.03} \\
\bottomrule
\end{tabular}
\end{table}

\noindent\textbf{Ablation studies.} We perform ablation studies to quantify the contributions of the proposed components i.e., slice-wise latent self-attention (SA) and confidence-aware curriculum learning (CL). We evaluate three variants, which include the baseline model without SA and CL, model with only SA, and the full model with both components activated. Tab.~\ref{tab:ablation} summarizes the results, demonstrating that SA improves overall segmentation performance, while the combination of SA and CL yields the best overall accuracy and scar volume error for difficult cases. These findings suggest SOTA methods underperform since they lack richer 3D context, operating on isolated 2D slices and random batch selection.

\begin{table}[htbp]
\centering
\setlength{\tabcolsep}{10pt}
\caption{Ablation study evaluating the contribution of our proposed components of SA and CL.}
\label{tab:ablation}
\begin{tabular}{lccccc}
\toprule
\multicolumn{2}{c}{Modules} & Dice $\uparrow$ & $\text{Dice}_\text{low}$ $\uparrow$ & SE (\%) $\downarrow$ & $\text{SE}_\text{low}$ (\%) $\downarrow$ \\
% \cmidrule(lr){1-2}
 {\scriptsize SA} & {\scriptsize SCL} & &  \\
\midrule
$\times$ & $\times$ & 0.571 $\pm$ 0.24 & 0.495 $\pm$ 0.27 & 53.68 & 50.34\\
$\checkmark$ & $\times$ & 0.639 $\pm$ 0.23 & 0.494 $\pm$ 0.29  & 47.46 & 46.89\\
$\times$ & $\checkmark$ & 0.644 $\pm$ 0.29 & 0.574 $\pm$ 0.23 & 45.07 & 48.34\\
$\checkmark$ & $\checkmark$ & \textbf{0.677 $\pm$ 0.24} & \textbf{0.644 $\pm$ 0.22} & \textbf{38.88}  & \textbf{35.03} \\
\bottomrule
\end{tabular}
\end{table}

\section{Conclusion}
In this paper, we present CalcSeg, a confidence-aware semi-supervised framework for myocardial scar segmentation from single-stack LGE-CMR. By defining an semi-supervised curriculum learning during the training process with an uncertainty-guided difficulty estimation and combining it with slice-wise attention for 3D latent context modeling, our method addresses the challenges of scar delineation in clinically challenging cases such as images with low contrast, heterogeneous scars and diffuse or small scar volume. Experiments on multi-center in-house and public datasets show improved robustness, particularly on challenging cases, highlighting the potential of CalcSeg as an annotation tool for automated clinical scar assessment. Future work will include integrating local-to-global curriculum approaches for network training, addressing uncertainty in scar tissue boundary, and incorporating multi-sequence CMR for better inflammation detection. \\

\noindent\textbf{Acknowledgements.} This work was supported by NIH 1R21EB032597, iPRIME Student Fellowship Award, and NSF CAREER 2239977.\\
\textbf{Disclosure of Interests.} The authors have no competing interests to declare that are
relevant to the content of this article.

\bibliographystyle{splncs04}
\bibliography{mybibfile}

\begin{thebibliography}{10}
\providecommand{\url}[1]{\texttt{#1}}
\providecommand{\urlprefix}{URL }
\providecommand{\doi}[1]{https://doi.org/#1}

\bibitem{arnab2021vivit}
Arnab, A., Dehghani, M., Heigold, G., Sun, C., Lu{\v{c}}i{\'c}, M., Schmid, C.: Vivit: A video vision transformer. In: Proceedings of the IEEE/CVF international conference on computer vision. pp. 6836--6846 (2021)

\bibitem{bengio2009curriculum}
Bengio, Y., Louradour, J., Collobert, R., et~al.: Curriculum learning. In: Proceedings of the 26th annual international conference on machine learning. pp. 41--48 (2009)

\bibitem{chen2021transunet}
Chen, J., Lu, Y., Yu, Q., et~al.: Transunet: Transformers make strong encoders for medical image segmentation. arXiv preprint arXiv:2102.04306  (2021)

\bibitem{coleman2017prognostic}
Coleman, G.C., Shaw, P.W., Balfour, P.C., et~al.: Prognostic value of myocardial scarring on cmr in patients with cardiac sarcoidosis. JACC: Cardiovascular Imaging  \textbf{10}(4),  411--420 (2017)

\bibitem{crozieruncertainty}
Crozier, T., Faure, A., Bos, D., et~al.: Uncertainty-based segmentation of myocardial infarction areas on cardiac mr images

\bibitem{cui2022deep}
Cui, H., Jiang, L., Yuwen, C., Xia, Y., Zhang, Y.: Deep u-net architecture with curriculum learning for myocardial pathology segmentation in multi-sequence cardiac magnetic resonance images. Knowledge-based systems  \textbf{249},  108942 (2022)

\bibitem{dice1945measures}
Dice, L.R.: Measures of the amount of ecologic association between species. Ecology  \textbf{26}(3),  297--302 (1945)

\bibitem{dosovitskiy2020image}
Dosovitskiy, A., Beyer, L., Kolesnikov, A., Weissenborn, D., Zhai, X., Unterthiner, T., Dehghani, M., Minderer, M., Heigold, G., Gelly, S., et~al.: An image is worth 16x16 words: Transformers for image recognition at scale. In: International Conference on Learning Representations (ICLR) (2021)

\bibitem{gal2016dropout}
Gal, Y., Ghahramani, Z.: Dropout as a bayesian approximation: Representing model uncertainty in deep learning. In: international conference on machine learning. pp. 1050--1059. PMLR (2016)

\bibitem{gao2025principled}
Gao, C., , et~al.: Principled data selection for alignment: The hidden risks of difficult examples. In: Proceedings of the 42nd International Conference on Machine Learning. Proceedings of Machine Learning Research, vol.~267, p. gao25f. PMLR (2025), \url{https://proceedings.mlr.press/v267/gao25f.html}

\bibitem{hasny2024myocardial}
Hasny, M., Demirel, O.B., Amyar, A., Faghihroohi, S., Nezafat, R.: Myocardial scar enhancement in lge cardiac mri using localized diffusion. In: International Conference on Medical Image Computing and Computer-Assisted Intervention. pp. 307--316. Springer (2024)

\bibitem{hatamizadeh2022unetr}
Hatamizadeh, A., Tang, Y., Nath, V., Yang, D., Myronenko, A., Landman, B., Roth, H.R., Xu, D.: Unetr: Transformers for 3d medical image segmentation. In: Proceedings of the IEEE/CVF winter conference on applications of computer vision. pp. 574--584 (2022)

\bibitem{he2025survey}
He, W., Jiang, Z., Xiao, T., Xu, Z., Li, Y.: A survey on uncertainty quantification methods for deep learning. ACM Computing Surveys  (2025)

\bibitem{hoh2024impact}
Hoh, T., Margolis, I., Weine, J., et~al.: Impact of late gadolinium enhancement image acquisition resolution on neural network based automatic scar segmentation. Journal of Cardiovascular Magnetic Resonance  \textbf{26}(1),  101031 (2024)

\bibitem{jayakumar2026curriculum}
Jayakumar, N., Pan, J., Wang, S., Paudel, B., Hosadurg, N., Singulane, C.C., Bhatt, S., Patel, A.R., Zhang, M.: Curriculum-guided myocardial scar segmentation for ischemic and non-ischemic cardiomyopathy. In: 2026 IEEE 23rd International Symposium on Biomedical Imaging (ISBI). pp.~1--5. IEEE (2026)

\bibitem{jayakumar2025deep}
Jayakumar, N., Pan, J., Wang, S., et~al.: Deep learning to predict myocardial scar burden and uncertainty quantification. Journal of Cardiovascular Magnetic Resonance  \textbf{27} (2025)

\bibitem{karim2016evaluation}
Karim, R., Bhagirath, P., Claus, P., et~al.: Evaluation of state-of-the-art segmentation algorithms for left ventricle infarct from late gadolinium enhancement mr images. Medical image analysis  \textbf{30},  95--107 (2016)

\bibitem{kellman2012cardiac}
Kellman, P., Arai, A.E.: Cardiac imaging techniques for physicians: late enhancement. Journal of magnetic resonance imaging  \textbf{36}(3),  529--542 (2012)

\bibitem{krueger2009flexible}
Krueger, K.A., Dayan, P.: Flexible shaping: How learning in small steps helps. Cognition  \textbf{110}(3),  380--394 (2009)

\bibitem{kwong2011measuring}
Kwong, R.Y., Farzaneh-Far, A.: Measuring myocardial scar by cmr (2011)

\bibitem{lalande2020emidec}
Lalande, A., Chen, Z., Decourselle, T., et~al.: Emidec: a database usable for the automatic evaluation of myocardial infarction from delayed-enhancement cardiac mri. Data  \textbf{5}(4), ~89 (2020)

\bibitem{lin2023continuation}
Lin, X., Yang, Z., Zhang, X., Zhang, Q.: Continuation path learning for homotopy optimization. In: International Conference on Machine Learning. pp. 21288--21311. PMLR (2023)

\bibitem{ma2024segment}
Ma, J., He, Y., Li, F., Han, L., You, C., Wang, B.: Segment anything in medical images. Nature Communications  \textbf{15}(1), ~654 (2024)

\bibitem{moafi2025robust}
Moafi, A., Moafi, D., Mirkes, E.M., et~al.: Robust deep learning for myocardial scar segmentation in cardiac mri with noisy labels. In: International Conference on Medical Image Computing and Computer-Assisted Intervention. pp. 529--539. Springer (2025)

\bibitem{oktay2018attention}
Oktay, O., Schlemper, J., et~al.: Attention u-net: Learning where to look for the pancreas. In: International Conference on Medical Imaging with Deep Learning (MIDL) (2018)

\bibitem{pathak2025principled}
Pathak, H.N., Paffenroth, R.: Principled curriculum learning using parameter continuation methods. arXiv preprint arXiv:2507.22089  (2025)

\bibitem{paudel2026cardiac}
Paudel, B., Hosadurg, N., Patel, R., et~al.: Cardiac mri parameters for risk prediction in cardiac sarcoidosis: Insights from a multicenter registry. Journal of Cardiovascular Magnetic Resonance  \textbf{28} (2026)

\bibitem{qiu2023myops}
Qiu, J., Li, L., Wang, S., et~al.: Myops-net: Myocardial pathology segmentation with flexible combination of multi-sequence cmr images. Medical image analysis  \textbf{84},  102694 (2023)

\bibitem{rajchl2015comparison}
Rajchl, M., Stirrat, J., Goubran, M., Yu, J., Scholl, D., Peters, T.M., White, J.A.: Comparison of semi-automated scar quantification techniques using high-resolution, 3-dimensional late-gadolinium-enhancement magnetic resonance imaging. The international journal of cardiovascular imaging  \textbf{31}(2),  349--357 (2015)

\bibitem{ronneberger2015u}
Ronneberger, O., Fischer, P., Brox, T.: U-net: Convolutional networks for biomedical image segmentation. In: International Conference on Medical image computing and computer-assisted intervention. pp. 234--241. Springer (2015)

\bibitem{tavakoli2025scarnet}
Tavakoli, N., Rahsepar, A.A., Benefield, B.C., Shen, D., L{\'o}pez-Tapia, S., Schiffers, F., Goldberger, J.J., Albert, C.M., Wu, E., Katsaggelos, A.K., et~al.: Scarnet: a novel foundation model for automated myocardial scar quantification from late gadolinium-enhancement images. Journal of Cardiovascular Magnetic Resonance p. 101945 (2025)

\bibitem{wang2022ai}
Wang, S., Patel, H., Miller, T., et~al.: Ai based cmr assessment of biventricular function: clinical significance of intervendor variability and measurement errors. Cardiovascular Imaging  \textbf{15}(3),  413--427 (2022)

\bibitem{xing2023joint}
Xing, J., Wang, S., Bilchick, K.C., et~al.: Joint deep learning for improved myocardial scar detection from cardiac mri. In: 2023 IEEE 20th international symposium on biomedical imaging (ISBI). pp.~1--5. IEEE (2023)

\bibitem{yang2025progressive}
Yang, B., Tian, Q., Liao, H., et~al.: Progressive curriculum learning with scale-enhanced u-net for continuous airway segmentation. BMC Medical Imaging  \textbf{26}, ~43 (2025)

\bibitem{zhang2020cascaded}
Zhang, Y.: Cascaded convolutional neural network for automatic myocardial infarction segmentation from delayed-enhancement cardiac mri. In: International Workshop on Statistical Atlases and Computational Models of the Heart. pp. 328--333. Springer (2020)

\bibitem{zheng2024curriculum}
Zheng, X., Zhang, Y., Zhang, H., et~al.: Curriculum prompting foundation models for medical image segmentation. In: International conference on medical image computing and computer-assisted intervention. pp. 487--497. Springer (2024)

\bibitem{zhou2020robust}
Zhou, T., Wang, S., Bilmes, J.: Robust curriculum learning: from clean label detection to noisy label self-correction. In: International conference on learning representations (2021)

\bibitem{zou2025scarelastic}
Zou, R., Li, Y., Zhou, N., Guan, X., Li, G., Xie, D., Bo, J., Li, S., Wu, M., Deng, J., et~al.: Scarelastic: continuous elasticity field modeling for myocardial scar delineation in lge-cmr. npj Digital Medicine  (2025)

\end{thebibliography}

\end{document}